\documentclass[12pt]{article}

\usepackage{natbib}

\usepackage{amssymb}
\usepackage{amsmath}
\usepackage{amscd}
\usepackage{latexsym}

\newcommand{\be}{\begin{equation}}
\newcommand{\ee}{\end{equation}}

\newcommand{\dlt}{\delta}

\newcommand{\bt}{\beta}
\newcommand{\rmM}{{\rm M}}

\newcommand{\al}{\alpha}
\newcommand{\ra}{\rightarrow}
\newcommand{\sgm}{\sigma}

\begin{document}

\begin{center}

{\Large {\bf Mathematical Structure of Quantum Decision Theory} \\ [5mm]

V.I. Yukalov$^{1,2}$ and D. Sornette$^{1,3}$ } \\ [3mm]

{\it $^1$Department of Management, Technology and Economics, \\
ETH Z\"urich, Z\"urich CH-8032, Switzerland \\ [3mm]

$^2$Bogolubov Laboratory of Theoretical Physics, \\
Joint Institute for Nuclear Research, Dubna 141980, Russia \\ [3mm]

$^3$ Swiss Finance Institute, \\
c/o University of Geneva, 40 blvd. Du Pont d'Arve,
CH 1211 Geneva 4, Switzerland }

\end{center}

\vskip 3cm

\begin{abstract}

One of the most complex systems is the human brain whose formalized
functioning is characterized by decision theory. We present a
``Quantum Decision Theory" of decision making, based on the
mathematical theory of separable Hilbert spaces. This mathematical
structure captures the effect of superposition of composite
prospects, including many incorporated intentions, which allows us
to explain a variety of interesting fallacies and anomalies that
have been reported to particularize the decision making of real human
beings. The theory describes entangled decision making,
non-commutativity of subsequent decisions, and intention
interference of composite prospects. We demonstrate how the
violation of the Savage's sure-thing principle (disjunction effect)
can be explained as a result of the interference of intentions, when
making decisions under uncertainty. The conjunction fallacy is also
explained by the presence of the interference terms. We demonstrate
that all known anomalies and paradoxes, documented in the context of
classical decision theory, are reducible to just a few mathematical
archetypes, all of which finding straightforward explanations in
the frame of the developed quantum approach.

\end{abstract}

\vskip 1cm


\vskip 5mm


\newpage

\section{Introduction}

The human brain is certainly one of the most complex systems known in nature
and it is therefore attracting strong interest in the attempts of understanding
its functioning. There are two ways of describing the latter. One way is
the study of physiological processes occurring in the brain, related to
physical and chemical effects governing the process of thinking. Another
approach is devoted to the formalized description of the thinking process
in terms of decision making theory. The latter approach forms the basis of
decision theory employed in social and economic sciences.

The scope of the present paper is the description of the brain functioning
in terms of decision theory. We propose a novel theory of decision making,
based on the mathematical theory of Hilbert spaces (Dieudonn\'e, 2006) and
employing the mathematical techniques that are used in the quantum theory
of physical measurements. Because of the latter, we call this approach
the {\it Quantum Decision Theory} (QDT). This approach can be thought of as
the mathematically simplest and most natural extension of objective
probabilities into nonlinear subjective probabilities. The proposed
formalism allows us to explain {\it quantitatively} without adjustable
parameters the known paradoxes, such as the disjunction and conjunction
effects. The disjunction effect is the failure of humans to obey the
sure-thing principle of classical probability theory. The conjunction
effect is a logical fallacy that occurs when people assume that specific
conditions are more probable than a single general one.  Our QDT also
unearths a deep relationship between the conjunction and the disjunction
effects. We show that the former is sufficient for the later to exist.

Decision theory is concerned with identifying what are the optimal
decisions and how to reach them. Traditionally, it is a part of
discrete mathematics. Much of decision theory is normative and
prescriptive, and assumes that people are fully-informed and
rational  \citep{9,49}. These assumptions have been questioned early
on with the evidence provided by the Allais paradox (Allais, 1953)
and by many other behavioral paradoxes (Camerer et al., 2003),
showing that humans often seem to deviate from the prescription of
rational decision theory due to cognitive and emotion biases.
Problems, occurring in the interpretation of classical utility
theory and its application to real human decision processes, have
been discussed in an extensive literature (e.g., Berger, 1985;
Weidlich, 1991; Zeckhauser, 2006; Machina, 2008). The theories of
bounded rationality (Simon, 1955) and of behavioral economics
\citep{47,Shefrin05} have been developed to account for these
deviations. As reviewed by Machina (2008), alternative models of
preferences over objectively or subjectively uncertain prospects
have attempted to accommodate these systematic departures from the
expected utility model, while retaining as much of its analytical
power as possible. In particular, non-additive nonlinear probability
models have been developed to account for the deviations from
objective to subjective probabilities observed in human agents
(Quiggin, 1982; Gilboa, 1987; Schmeidler, 1989; Gilboa and
Schmeidler, 1989; Cohen and Tallon, 2000; Montesano, 2008). Decision
making in the presence of uncertainty about the states of nature has
been formalized in the statistical decision theory (Lindgren, 1971;
White, 1976; Hastings and Mello, 1978; Rivett, 1980; Buchanan, 1982;
Berger, 1985; Marshall and Oliver, 1995; Bather, 2000; French and
Insua, 2000; Raiffa and Schlaifer, 2000; Weirich, 2001). However,
many paradoxes remain unexplained or are sometimes rationalized on
an ad hoc basis, which does not provide much predictive power.

To break this stalemate, our approach relies and extends \cite{5}
theory of quantum measurements in two principal aspects. First of all, we
generalize the theory of quantum measurements to {\it active} objects, which
are not simply passive systems exposed by a measurement performed by an
external observer, but which are complex decision makers themselves,
characterized by their own strategic state. Certainly, humans are extremely
complex systems with many inaccessible features, such as emotions, biases,
and subconscious processes. Indeed, even the most modern and precise
measurements and localization of brain activity offer at best a rough
cartography of physical operations only indirectly related to the thought
processes. While inaccessible, one should not underestimate the importance
of emotions and of subconscious processes in decision making, as is
evidenced by a growing body of multidisciplinary research. The problem is
that these inaccessible features of brain processes cannot be accounted for
in the decision procedures based on the classical utility theory \citep{9}
and its extensions. As a consequence, a variety of different anomalies and
paradoxes, some of them already mentioned, have emerged from the
confrontation of classical decision theory with empirical data on the
decisions and choices made by real humans. In contrast, QDT is based on a
minimalist set of axioms that lead to predictions on decision making that
are in agreement with observed behaviors.

While being a direct descendent of the \cite{5} theory of measurement, our
approach is more general, complementing it by novel notions that are,
necessary for practical applications to active decision makers. The second
important point of our theory is that, beyond simple actions, we
consider {\it complex} actions that can be composed of several action
representations or action modes. The relevance of the quantum description
for classical objects can be justified by the following analogy. Quantum
theory, as is known, could be interpreted as a cross-section of an
underlying classical theory equipped with a large number of {\it contextual
hidden variables} (see, e.g., Yukalov, 1975). However, in order to describe
the results of $N$ measurements on a single quantum system, it is necessary
to involve a number of hidden variables which is proportional to $N$
(Dakic et al., 2008). Hence, to describe any number of experiments on many
systems, it is necessary to introduce an infinite number of hidden variables.
Such a way of describing physical measurements is, evidently, unpractical.
The techniques of quantum theory, avoiding the necessity of dealing with
numerous unspecified hidden variables, provide the simplest method of
characterizing any number of physical measurements on an arbitrary number
of systems. In the case of decision theory, the role of hidden variables
is played by various biases, underlying emotions, subconscious feelings,
and other hidden features characterizing decision makers. This is why the
quantum approach seems to provide the simplest way of taking into account
such hidden characteristics, at the same time avoiding the necessity of
explicitly dealing with them, similarly to avoiding the hidden variables
in quantum theory.

Our approach is based on more than just the argument that human decisions
involve complex processes beyond the reach of detailed empirical or
theoretical analysis. It is based on the hypothesis that the thought
processes, involved in the definition and analysis of alternative
prospects and scenarios, do not necessarily separate them according to the
recipes of standard probability theory. Our QDT formalizes systematically
a broader class of decision making processes in which prospects can interact,
interfere and remain entangled. Our formulation of  QDT thus captures the
effect of superposition of composite prospects, including many incorporated
intentions. Dealing with composite prospects involves entangling operations
and the appearance of decision interference, characterizing decision
procedures under the perception of uncertainty and/or potential harmful
consequences. When such a generalization to standard probability theory and
to the usual optimization of classical decision theory is introduced,
we find that all the paradoxes are explained as unavoidable consequences of
the more general process of decision making described by the QDT. We apply
the developed machinery to several examples typical of various paradoxes of
classical decision theory. By the corresponding theorems, we prove that no
such paradoxes exist in QDT.

Historically, the idea that the process of human thinking can be described
by the techniques of quantum theory was advanced by Bohr (1933, 1937).
Using these techniques, it would be possible to characterize the complex
psychological processes occurring in human mind and accompanied by the
effects typical of quantum theory, for instance, by interference. From
time to time, the Bohr ideas have been revived in the literature
(e.g., Busemeyer et al., 2006). However, no decision theory has been
suggested, which would have predictive power.

While developing the mathematical theory of quantum measurements,
Von Neumann (1955) mentioned that the measurement procedure, to some
extent, reminds of the process of taking decisions. This analogy has been
further emphasized by Benioff (1972) in his description of quantum
information processing. The basics of a general approach, treating human
decision making as a kind of quantum theory of measurements, was announced
in the brief letter (Yukalov and Sornette, 2008). The possibility of
constructing a thinking quantum system, imitating the process of human
decision making, has been advanced (Yukalov and Sornette, 2009). It was
also shown that the QDT approach provides a natural explanation for the
phenomenon of dynamic inconsistency (Yukalov and Sornette, 2009). The main
aim of the present paper is twofold. First, we develop the general Quantum
Decision Theory, describing in detail the basic mathematical structure of
QDT. Second, we prove several theorems explaining the origin of the
classical paradoxes in standard decision making and show that such
paradoxes do not appear in QDT. The rigorous mathematical proof that QDT is
free of any kind of the paradoxes, plaguing classical decision theory, is
the principal novel result of this paper.

In our recent paper (Yukalov and Sornette, 2009), we have concentrated
our attention on the phenomenon of dynamic inconsistency, which we do not
touch here. In contrast, we analyze here other paradoxes that have not
been treated in our previous papers, which makes the main content of the
present paper principally different and new.

It is also worth stressing the main points distinguishing our approach
from the von Neumann quantum theory of measurement:

(i) First of all, in quantum measurements, a {\it passive} quantum system
is subject to a measuring action from an {\it external} observer. Contrary
to this, a decision maker is an {\it active} subject taking decisions by
himself/herself, thus, the decisions being the product of an {\it internal}
process.

(ii) The measuring devices, acting on quantum systems, are more or less
the same for different systems. Mathematically, the measurement procedure
can be described by calculating the expectation values for local observables,
with respect to any complete basis. But a decision maker is a personality
that is characterized by his/her own features, which mathematically implies
that there exists a specific {\it strategic state} associated with this
decision maker. Therefore any expectation value is to be taken with respect
to this strategic state, and not with respect to an arbitrary basis. This
point is one of the principal differences with the previous attempts
to use the formalism of quantum mechanics to decision theory.

(iii) Quantum systems, in the majority of cases, are relatively simple, as
compared to human brains. In the process of decision making, one usually
considers not just simple actions but rather complicated prospects, composed
of several actions. This makes it necessary to systematically consider
composite prospects, which is somewhat similar to dealing with complex
systems with a complicated internal structure.

The organization of the paper is as follows. In order to emphasize that
the theory is problem-driven, we present in Sec. 2 several most often
cited paradoxes in classical decision making, the Allais paradox, the
Ellsberg paradox, and the Kahneman-Tversky paradox. Section 3
introduces the definition of ``intended actions'' and ``prospects'' and
of the mathematical notions to describe them. Section 4 defines the
mathematical structure of the mode states, mode space, and the space of
mind. Section 5 introduces the prospect probability operators. Section 6
defines the prospect probabilities as being the observables in the
decision process. Section 7 formulates the rules according to which a
decision is performed within the mathematical construction of QDT.
Section 7 also describes the phenomenon of decision interference occurring
under uncertainty and/or perceived potential harm. For this, the notion of
composite prospects is introduced and the principle of ``aversion to
uncertainty and loss'' is defined. Section 8 applies the previous
definitions and concepts to binary decision cases, as they form the most
common situation in the empirical literature. Section 9 presents the major
results of the theory, showing how the main paradoxes documented in the
literature are explained within QDT. The following paradoxes are addressed:
the Allais paradox (or compatibility paradox), the independence paradox,
the Ellsberg paradox, the inversion paradox, the invariance violation,
described by Kahneman and Tversky, the certainty effects, the disjunction
effect (or violation of the Savage sure-thing principle), the conjunction
fallacy, and the isolation effect (or focusing, availability, salience,
framing, or elicitation effects). Section 9 ends with a prediction on the
condition under which several of these paradoxes can occur simultaneously.
In particular, we predict that the conjunction fallacy should be accompanied
by the disjunction effect. Section 10 summarizes and concludes.

\section{Some Famous Paradoxes}

There exist several paradoxes in classical decision theory based on the
notion of expected utility. Despite numerous attempts to resolve these
paradoxes, none of the suggested approaches have been able to resolve
all of them in the frame of the same theory. Various extensions of utility
theory, based on constructing non-expected utility functionals
(Machina, 2008), do not resolve these paradoxes, as has been proved by Safra
and Segal (2008). Usually, extending the classical utility theory ``ends up
creating more paradoxes and inconsistencies than it resolves" (Al-Najjar
and Weinstein, 2009). In order to stress that the necessity of advancing
a novel variant of decision theory, the QDT presented here, is not just
``theory-driven" but is fundamentally ``problem-driven", with the aim of
resolving the existing paradoxes, we describe below some of the most often
discussed paradoxes occurring in classical decision making.

\subsection{Allais paradox}

The Allais paradox expresses the observation that several decisions which
are mutually incompatible, according to classical utility theory, are
nevertheless embraced by real human people \citep{44}. The mathematical
structure of this paradox is as follows. One considers four alternatives
corresponding to four prospects. For example, let us consider the
actions $A_n$ corresponding to the $n$-th choice procedure (prospect), with
$n$ taking the values $1, 2, 3$ and $4$. Let the $X_j$'s
be the amounts of money to be gained with the corresponding probabilities
$p_n(X_j)$. The utility function, measuring the utility of getting an
amount $X_j$ of money is $U(X_j)$. According to utility theory, a prospect
$\pi_n$ is characterized by a lottery $\{ p_n(X_j),U(X_j)\}$. The expected
utility of the prospect $\pi_n$ is
\be
\label{A1}
U(\pi_n) = \sum_j \; p_n(X_j) U(X_j) \; .
\ee
One also assumes the balance condition among the four prospects:
\be
\label{A2}
p_1(X_j) + p_3(X_j) = p_2(X_j) + p_4(X_j)   ~~~{\rm for ~all}~ j\; .
\ee
In the Allais paradox, subjects prefer $\pi_1$ to $\pi_2$ which, according
to utility theory, implies that
\be
\label{A3}
U(\pi_1) \; > \; U(\pi_2) \; .
\ee
And $\pi_3$ is preferred or indifferent to $\pi_4$, which means that
\be
\label{A4}
U(\pi_3) \; \geq \; U(\pi_4) \; .
\ee
Using the definition of the expected utility (\ref{A1}), one has from
(\ref{A3}) the equation
\be
\label{A5}
\sum_j \; [ p_1(X_j) - p_2(X_j) ] U(X_j) \; > \; 0 \; ,
\ee
while from Eq. (\ref{A4}) one gets
\be
\label{A6}
\sum_j \; [ p_3(X_j) - p_4(X_j) ] U(X_j) \; \geq \; 0 \; .
\ee
Using the balance condition (\ref{A2}), the latter inequality transforms
into
\be
\label{A7}
\sum_j \; [ p_1(X_j) - p_2(X_j) ] U(X_j) \; \leq \; 0 \; .
\ee
Equations (\ref{A5}) and (\ref{A7}) are evidently incompatible.

In the original Allais paradox, the amounts of money were chosen so that
$U(X_1)<U(X_2)<U(X_3)$, with $X_1\ra\$ 0$, $X_2\ra\$ 10^6$, and $X_3\ra\$
5\times 10^6$. The probabilities $p_n(X_j)$ were defined as
\be
\{ p_1(X_j)\} = \{ 0,1,0\} \; , \qquad
\{ p_2(X_j)\} = \{ 0.01,0.89,0.10\} \; , \hspace{4.5cm}
\ee
\be
\label{A8}
\{ p_3(X_j)\} = \{ 0.90,0,0.10\} \; , \qquad
\{ p_4(X_j)\} = \{ 0.89,0.11,0\} \; .
\ee
The balance condition (\ref{A2}) here is valid, since
$$
p_1(X_1) + p_3(X_1) = p_2(X_1) + p_4(X_1) = 0.9 \; ,
$$
$$
p_1(X_2) + p_3(X_2) = p_2(X_2) + p_4(X_2) = 1 \; ,
$$
$$
p_1(X_3) + p_3(X_3) = p_2(X_3) + p_4(X_3) = 0.1 \; .
$$
Decisions, when the prospect $\pi_1$ is preferred to $\pi_2$ and the
prospect $\pi_3$ is preferred to $\pi_4$, are clearly incompatible.
The explanation of this paradox, in the framework of QDT, will be given
in Sec. 9, being based on the fact that the prospect $\pi_2$ is more
uncertain than $\pi_1$, while the prospects $\pi_3$ and $\pi_4$ are of
a comparable uncertainty.

The balance condition (\ref{A2}) also leads to the violation of the
independence axiom of expected utility theory. This axiom postulates
that if the prospect $\pi_1$ is strictly preferred to $\pi_2$, and the
prospect $\pi_3$ is preferred or indifferent to $\pi_4$, in the sense
of conditions (\ref{A3}) and (\ref{A4}), then the prospect $\pi_1+\pi_3$
is strictly preferred to $\pi_2+\pi_4$, so that
\be
\label{A9}
U(\pi_1 + \pi_3) \; > \; U(\pi_2 + \pi_4) \; .
\ee
However, under the balance condition (\ref{A2}), one has
\be
\label{A10}
U(\pi_1 + \pi_3) \; = \; U(\pi_2 + \pi_4) \; .
\ee
Equations (\ref{A9}) and (\ref{A10}) violate the independence axiom of
utility theory. This anomaly will be resolved in Proposition 6.

\subsection{Ellsberg paradox}

The Ellsberg paradox demonstrates that no utility function can describe
the choices made by real human agents for the following process
\citep{45}. One considers two prospects $\pi_1$ and $\pi_2$, defined as
the conjunctions $\pi_n = A_n X$, with $X = X_1 + X_2 + \ldots$. And one
assumes the validity of the equivalence condition
\be
\label{B1}
p_1(X_j) = p_2(X_j) \qquad (\forall j)\; .
\ee
Under this condition, the expected utility functions (\ref{A1}) for both
prospects $\pi_1$ and $\pi_2$ coincide,
\be
\label{B2}
U(\pi_1) = U(\pi_2) \; .
\ee
This equality holds for arbitrary utility functions. But in the
realization of this paradox, one defines the probabilities $p_1(X_j)$
explicitly, while keeping the value of $p_2(X_j)$ unknown or, perhaps
better said, ambiguous. This ambiguity does not remove the fact that
condition (\ref{B1}) holds true. But the value of $p_2(X_j)$ is perceived
less clearly or, using different framing of the decision making process,
$\pi_2$ is organized as a more uncertain prospect than $\pi_1$. It turns
out that subjects always prefer $\pi_1$ to $\pi_2$, which necessarily
implies
\be
\label{B3}
U(\pi_1) \; > \; U(\pi_2) \; ,
\ee
if classical utility theory is to describe this situation. But, Eqs.
(\ref{B2}) and (\ref{B3}) are in contradiction with each other, which
implies that there are no utility functions that would satisfy both these
equations simultaneously.

Such a paradox does not arise in QDT, as will be  proved in Proposition 7.

\subsection{Kahneman-Tversky paradox}

The logical meaning of the Kahneman-Tversky paradox is as follows \citep{47}.
One considers four prospects, $\pi_n = A_n X$, with $n = 1,2,3,4$ and the
structure is as in the Ellsberg paradox above. The actions $A_n$ characterize
the choices, enumerated with $n=1,2,3,4$. The alternatives $X_j$ describe
the amounts of money which are such that $U(X_1)<U(X_2)<U(X_3)$ and
\be
\label{C1}
U(X_2) = \frac{1}{2}\left[ U(X_1) + U(X_3) \right]\; .
\ee
The probabilities $p_n(X_j)$ are given in such a way that
\be
\label{C2}
U(\pi_n)\; = \; const \qquad (n=1,2,3,4) \; ,
\ee
with the expected utility function defined as in Eq. (\ref{A1}).

The procedure is organized so that the prospect $\pi_1$ provides a more
uncertain gain than $\pi_2$, hence, $\pi_1$ is more uncertain that $\pi_2$.
And the prospect $\pi_4$ is defined as yielding more certain loss than
$\pi_3$, consequently, $\pi_4$ is more repulsive than $\pi_3$. In response
to the involved risks, real human subjects prefer $\pi_2$ to $\pi_1$ and
$\pi_3$ to $\pi_4$, which implies
\be
\label{C3}
U(\pi_1)\; < \; U(\pi_2) \; , \qquad U(\pi_3)\; > \; U(\pi_4) \; ,
\ee
in contradiction with Eq. (\ref{C2}).

In the original Kahneman-Tversky paradox \citep{42}, the probability sets
were defined as
$$
\{ p_1(X_j) \} = \{ 0.5,0,0.5\} = \{ p_3(X_j)\} \; , \qquad
\{ p_2(X_j) \} = \{ 0,1,0\} = \{ p_4(X_j)\} \; .
$$

The Kahneman-Tversky paradox does not occur in QDT, as will be proven in
Proposition 9.

\section{Intended Actions and Prospects}

Decision making is concerned with the choice between several intended
actions that, for brevity, can be called intentions or just actions.
The analog of actions in classical probability theory is the notion
of events forming a field of events. In contrast, in QDT, actions are
not necessarily commutative and, generally, do not compose a field.
Noncommutativity of actions in QDT is similar to the noncommutativity
of events in noncommutative probability theory \citep{16}. This echoes
the general observation that intended actions in real life are also
very often noncommutative. It is interesting to note that Pierce (1880)
was, probably, the first to emphasize that logical statements are
generally noncommutative. The formalization of his relative logic
\citep{17} is based on the recognition that the order of logical
statements can be essential. We start from these premises to describe
and analyze the algebra of actions within QDT. In the Introduction, we
have described the main ideas of QDT in simple words. In what
follows, we turn to a more rigorous consideration in the style widely
accepted in economic literature and in the literature on mathematics of
complex systems.

\subsection{Action ring}

Let us consider a set of intended actions
\be
\label{eq1}
{\cal A} =  \{ A_n: \; n=1,2,\ldots \} \; ,
\ee
where the actions are enumerated with the index $n=1,2,\ldots$. It is
natural to require the validity of the binary operation {\it addition},
such that for every $A,B\in{\cal A}$ their sum $A+B\in{\cal A}$. The sum
$A+B$ means that either the action $A$ or action $B$, or both, are intended
to be accomplished, hence $A+B=B+A$. The addition is assumed to be
associative, so that for any $A,B,C\in{\cal A}$, one has $A+(B+C)=(A+B)+C$.
By direct extension, the addition is defined as a reversible operation,
such that from $A+B=C$ it follows that $A=C-B$. Thus, the elements of the
action set (\ref{eq1}) form the action group, which is an Abel group with
respect to addition.

Another natural binary operation for the action of set (\ref{eq1}) is {\it
multiplication}, when for any $A,B\in{\cal A}$, there exists $AB\in{\cal A}$.
The product of actions $AB$ implies that both actions are intended to occur
together. The multiplication can be defined as a distributive operation, for
which $A(B+C)=AB+AC$. However, since the order of the actions can be
important, the multiplication, in general, is not commutative, so that $AB$
is not necessarily the same as $BA$. When writing $AB$, we mean that the
action $B$ is to be accomplished before $A$. This makes it painless to define
the associative triple product $ABC=(AB)C=A(BC)$. The zero element $0$ can
be introduced as that satisfying the equalities $A\cdot 0=0\cdot A=0$
for any $A\in{\cal A}$. This element characterizes an {\it empty action},
that is, an impossible action. If one would assume that the action set
(\ref{eq1}) does not contain the divisors of zero, so that $AB\neq 0$ for
any nonzero $A$ and $B$, then the nonzero elements of set (\ref{eq1}) would
form a groupoid with respect to multiplication. However, this is not the
case as the action set (\ref{eq1}) does indeed contain divisors of zero:
any product of two disjoint actions, by definition, is an empty action,
so that mutually disjoint actions are divisors of zero with respect to each
other. By this definition, two nonzero actions are disjoint if and only if
their intersection is an empty action.

The described properties characterize the action set (\ref{eq1}) as a
noncommutative ring, which will be called the {\it action ring}. A set
of subsets of ${\cal A}$, closed with respect to countable unions and
complementations, is a $\sgm$-ring, and if ${\cal A}$ pertains to this
set, it is a $\sgm$-field. For the purpose of developing our QDT, we need
to specify some constructions that can be formed with the actions
belonging to the action ring ${\cal A}$.

\subsection{Action modes}

In decision theory, an intended action can possess several
representations corresponding to different particular ways of realizing
this action. Such a {\it composite action} takes the form of
the union
\be
\label{eq2}
A_n \equiv \bigcup_{\mu=1}^{M_n} A_{n\mu} \qquad (A_{n\mu} \in{\cal A} ) \; ,
\ee
whose partial terms are the {\it action modes}, with $M_n\geq 1$ being
the number of modes. When $M_n=1$, the action (\ref{eq2}) is simple.

\subsection{Action prospects}

A more complicated structure is an {\it action prospect}, which is an
intersection
\be
\label{eq3}
\pi_n \equiv \bigcap_j A_{n_j} \qquad
(A_{n_j} \subset {\cal A}_n \subset {\cal A} )
\ee
of the actions $A_{n_j}$ from a subset ${\cal A}_n$ of the action ring
${\cal A}$. The actions in the product (\ref{eq3}) can be either simple
or composite as in the union (\ref{eq2}). The difference between $A_{n_J}$
in (\ref{eq3}) and $A_{n\mu}$ in (\ref{eq2}) is that the former can be a
composite action, while the latter is a simple action representing a single
action mode.

\subsection{Elementary prospects}

The simplest structure among the action prospects (\ref{eq3}) corresponds
to the prospects including only the simple single-mode actions or separate
modes of composite actions. Let $\{ j_n\}_\al$ be a set of indices labelling
simple actions or separate modes. The {\it elementary prospect} is the
product
\be
\label{eq4}
e_\al \equiv \bigcap_n A_{n j_n} \qquad (j_n \in \{ j_n\}_\al ) \; ,
\ee
where each $A_{n j_n}$ is simple and enters only one of the prospects,
so that different elementary prospects are disjoint in the sense that
their conjunction is an empty action, $e_\al e_\bt=0\;(\al\neq\bt)$.

\subsection{Prospect lattice}

For use as a decision theory, it is important that the set of all possible
prospects could be in some sense ordered. Suppose that the set $\{\pi_n\}$
of admissible prospects can be organized so that, for each two prospects from
$\{\pi_n\}$, an ordering binary relation $\leq$ can be defined. Then
for each pair $\pi_1$ and $\pi_2$, one should have either $\pi_1\leq\pi_2$ or
$\pi_2\leq\pi_1$. The ordering relation is linear, such that $\pi_1\leq\pi_2$
implies $\pi_2\geq\pi_1$. And it is transitive, so that, if $\pi_1\leq\pi_2$
and $\pi_2\leq\pi_3$, then $\pi_1\leq\pi_3$. Such a partially ordered
manifold $\{\pi_n\}$ composes a {\it lattice}.

The lattice is assumed to be {\it complete}, containing the minimal and the
maximal elements, for which
\be
\label{eq5}
\inf_n \pi_n \equiv 0 \; , \qquad \sup_n \pi_n \equiv \pi_* \; .
\ee
Thus, we obtain a complete lattice of partially ordered prospects,
\be
\label{eq6}
{\cal L} \equiv \{ \pi_n : \; 0\leq \pi_n \leq \pi_* \} \; ,
\ee
named the {\it prospect lattice}. An explicit ordering procedure will be
defined below.

\section{Mode and Prospect States}

The notion of states is crucial in quantum theory \citep{18}. And it is
equally important in QDT. In order to represent the quantum-mechanical
states, we shall employ the Dirac notation \citep{19,20}, which allows for
a very compact representation of formulas.

\subsection{Mode states}

For every single mode $A_{nj}$ of an action $A_n$, we put into correspondence
a state $|A_{nj}>$, which is a function ${\cal A}\ra\mathbb{C}$. A conjugate
state (in the sense of a Hermitian conjugate) is denoted as $<A_{nj}|$, and
will be necessary in the definition of the scalar product and for introducing
the probability operators associated with intentions and prospects. For each
two mode states, a scalar product is defined as $<A_{ni}|A_{nj}>$, which is
a function ${\cal A}\times{\cal A}\ra\mathbb{C}$. The mode states can be
normalized to one, and they are assumed to be orthogonal, so that
$$
< A_{ni}|A_{nj} > \; = \; \dlt_{ij} \; ,
$$
where $\dlt_{ij}$ is the Kronecker symbol. The orthogonality here reflects
the fact that the action modes $A_{ni}$ and $A_{nj}$ for $i\neq j$ represent
incompatible actions.

\subsection{Mode space}

The closed linear envelope, spanning all mode states, composes the mode
space
\be
\label{eq7}
{\cal M}_n \equiv {\rm Span} \{ | A_{nj} > : \; j=1,2,\ldots,M_n \} \; .
\ee
The dimensionality of ${\cal M}_n$ is ${\rm dim}{\cal M}_n = M_n$. For
each two elements $|A>$ and $|B>$ of ${\cal M}$, a scalar product is defined
enjoying the property $<A|B>=<B|A>^*$, where the symbol $*$ denotes
the complex conjugate, and possessing all other properties characterizing
scalar products. That is, the mode space (\ref{eq7}) is a Hilbert space.

\subsection{Basic states}

Using the mode states $|A_{nj}>$, it is possible to construct the states
of the elementary prospects (\ref{eq4}) as functions
${\cal A}\times{\cal A}\times\ldots \times{\cal A}\ra\mathbb{C}$,
\be
\label{eq8}
| e_\al> \; \equiv \; | A_{1j_1}  A_{2j_2} \ldots > \; \equiv \;
\bigotimes_n |  A_{nj_n} > \; ,
\ee
where $j_n\in\{ j_n\}_\al$. The states (\ref{eq8}) are termed the basic
states. Their scalar product is given as
$$
< e_\al | e_\bt> \; = \; \prod_n < A_{ni_n} | A_{nj_n} > \; ,
$$
where $i_n\in\{ i_n\}_\al$ and $j_n\in\{ j_n\}_\bt$. Given the
orthonormality of the mode states, one has
$$
< e_\al | e_\bt> \; = \; \dlt_{\al\bt} \; .
$$
The orthonormality of the basic states again reflects the fact that two
elementary prospects $e_\al$ and $e_\bt$ are incompatible for $\al\neq\bt$.

\subsection{Space of mind}

The space of mind, or {\it mind space}, is the closed linear envelope
\be
\label{eq9}
{\cal M} \equiv {\rm Span} \{ | e_\al> \} =
\bigotimes_n {\cal M}_n \; ,
\ee
spanning all basic states (\ref{eq8}). Its dimensionality is
\be
\label{eq10}
{\rm dim} {\cal M} = {\displaystyle \prod_n {\rm dim} {\cal M}_n } \; .
\ee
The space of mind (\ref{eq9}) is a Hilbert space, represented as a tensor
product of the Hilbert spaces ${\cal M}_n$. If the number of Hilbert spaces
${\cal M}_n$ is infinite, the space (\ref{eq9}) is understood as a von
Neumann infinite tensor product \citep{21}.

\subsection{Strategic state of mind}

The strategic state of mind $|\psi_s>$ is a normalized vector of ${\cal M}$.
As any vector of the Hilbert space (\ref{eq9}), it can be expanded over the
basic states (\ref{eq8}),
\be
\label{eq11}
|\psi_s> \; = \; \sum_\al \; c(e_\al)| e_\al> \; .
\ee
The normalization condition reads as
\be
\label{eq12}
< \psi_s|\psi_s> \; = \; \sum_\al \; |c(e_\al)|^2 = 1 \; .
\ee
The strategic state of mind characterizes the mind of the considered
decision maker, describing the latter as a particular being, with his/her
specific features as well as with the concrete available information
(Dixit and Besley, 1997). Each decision maker possesses his/her own
strategic state. We stress that this strategic state of mind $|\psi_s>$
is specific of a person at a given time and may display temporal evolution,
according to different homeostatic processes adjusting the individual to
the changing environment (Dawkins, 2006).

\subsection{Prospect states}

To each action prospect (\ref{eq3}), one can put into correspondence a
prospect state $|\pi_n>\;\in{\cal M}$. As any member of ${\cal M}$, it
can be expanded over the basic states (\ref{eq8}),
\be
\label{eq13}
|\pi_n> \; = \; \sum_\al \; b_n(e_\al)| e_\al> \; .
\ee
The states (\ref{eq13}) do not need to be neither orthogonal nor
normalized, which assumes that different prospects are not necessarily
disjoint, but could be realized together in the probabilistic sense
described below.

For the empty prospect, there corresponds the {\it vacuum state}, or
{\it empty state} $|0>$, such that
$$
< 0| e_\al> \; = \; < e_\al|0> \; = \; 0
$$
for any basic state $|e_\al>\;\in {\cal M}$.

\section{Probability Operator Measure}

In order to introduce a probability operator measure, we consider the set
of prospects forming a complete lattice. Specifying a prospect as a
composite action, we keep in mind that real decisions are usually made by
choosing among composite actions, but not among simple structureless actions.
As we show below, the existence of internal structures within prospects has
consequences of great importance for applications.

Let $\Sigma({\cal L})$ be a set of subsets of the prospect lattice ${\cal L}$,
closed under countable unions and complementations. Then $\Sigma({\cal L})$
is a $\sgm$-ring of ${\cal L}$. And let ${\cal B}({\cal M})$ be the set of
all bounded linear positive operators on the Hilbert space ${\cal M}$. Let us
consider a positive mapping $\hat P: \; \Sigma({\cal L})\ra{\cal B}({\cal M})$
enjoying the following properties. For an empty prospect,
\be
\label{eq14}
\hat P(0) = \hat 0 \; ,
\ee
where $\hat 0$ is the zero operator in ${\cal B}({\cal M})$. The mapping
$\hat P$ is self-adjoint, so that for any $\pi_n\in{\cal L}$, one has
\be
\label{eq15}
\hat P^+(\pi_n)  = \hat P(\pi_n) \; ,
\ee
where the cross implies the Hermitian conjugate. A mapping $\hat P$,
satisfying conditions (\ref{eq14}) and (\ref{eq15}), is an {\it operator
measure}.

Including ${\cal L}$ in $\Sigma({\cal L})$ makes the latter a $\sgm$-field.
Then $\hat P({\cal L})$ can be defined as a unity operator. In what follows,
we assume that it is a unity operator in the weak sense, with respect to the
matrix elements of the strategic state:
\be
\label{eq16}
<\psi_s| \hat P({\cal L}) |\psi_s> = 1 ~ .
\ee
An operator measure, satisfying conditions (\ref{eq14}), (\ref{eq15}),
and (\ref{eq16}), is the {\it probability operator measure}
$\{{\cal L},\Sigma({\cal L}),\hat P\}$. More information on operator
measures can be found, for instance, in the book \citep{22}.

In our case, the probability operator measure $\hat P$ is defined on the
$\sgm$-ring $\Sigma({\cal L})$ of the prospect lattice ${\cal L}=\{\pi_n\}$.
This probability operator measure includes the probability operators of
the prospects $\hat P(\pi_n)$. The latter can be represented in the
Dirac notation as
\be
\label{eq17}
\hat P(\pi_n) \equiv | \pi_n><\pi_n| \; .
\ee

The {\it prospect probability operators} (\ref{eq17}) are similar to
projection operators, but are not exactly projectors as they lack
the property of being idempotent. In addition, the operators (\ref{eq17})
are not commutative. But it is known \citep{6,7,8}  that the operators of
such measures do not need to be neither projection operators nor have to
commute with each other.

The set of all probability operators (\ref{eq17}), over the field of
complex numbers $\mathbb{C}$, forms the {\it algebra of probability
operators}
\be
\label{eq18}
{\cal P} \equiv \left \{ \hat P(\pi_n): \; \pi_n\in {\cal L}
\right \} \; .
\ee
The involution in ${\cal P}$ is defined as the Hermitian conjugation.
Since operators (\ref{eq17}) are self-adjoint, the involution in ${\cal P}$
is a bijection. Thus, ${\cal P}$ is an involutive bijective algebra.

A property of the probability operators (\ref{eq17}), which is of
importance for QDT, is that $\hat P(\pi_n)$ is, generally, an entangling
operator. One should not confuse entangled states and entangling operators.
Entangled states are those that cannot be represented as tensor products
of partial states \citep{23,24,25}. Entangling operations are those that
produce entangled states, even when acting on disentangled states
\citep{26,27,28}. Entangling operations in decision theory are of great
importance, resulting in a variety of nontrivial effects. The entangling
property of the probability operators (\ref{eq17}) is due to the composite
structure of the action prospects $\pi_n$.

\section{Prospect Operators as Observables}

The algebra of the probability operators (\ref{eq18}) is analogous to the
algebra of local observables in quantum theory \citep{29,30,31}. In quantum
theory, observable quantities are defined as the expectation values, or
averages, of the operators over the algebra of local observables. The
complete set of these expectations composes the statistical state of a
quantum system.

The observable quantities in QDT are the expectation values of the
probability operators, which define the scalar probability measure.
The expectation value of a probability operator $\hat P(\pi_n)$, under
the strategic state of mind $|\psi_s>$, is
\be
\label{eq19}
< \hat P(\pi_n)> \; \equiv \; < \psi_s | \hat P(\pi_n) | \psi_s > \; .
\ee
The corresponding observable quantity is the {\it prospect probability}
\be
\label{eq20}
p(\pi_n) \; \equiv \; < \hat P(\pi_n) > \; .
\ee
Taking into account the explicit form of the probability operator
(\ref{eq17}), from Eqs. (\ref{eq19}) and (\ref{eq20}), we have
\be
\label{eq21}
p(\pi_n) = | < \pi_n|\psi_s> |^2 \; .
\ee
This expression shows that $p(\pi_n)$ is evidently non-negative.
Of course, the normalization condition
\be
\label{eq22}
\sum_n \; p(\pi_n) =  1
\ee
is imposed, where the summation is over all $\pi_n\in{\cal L}$. Thus,
the mapping (\ref{eq20}) defines the scalar probability measure
$p:\;{\cal L}\ra[0,1]$.

\vskip 2mm

{\bf Definition 6.1}. {\it The probability $p(\pi_n)$ of a prospect
$\pi_n\in{\cal L}$ is the expectation value (\ref{eq20}) of the
probability operator (\ref{eq17}), with normalization condition
(\ref{eq22})}.

\vskip 2mm

The scalar probability measure, introduced above, makes it possible to
realize the ordering of the prospects in the prospect lattice ${\cal L}$,
classifying them as follows.

\vskip 2mm

{\bf Definition 6.2}. {\it Two prospects, $\pi_1$ and $\pi_2$ from the
prospect lattice ${\cal L}$, are indifferent $(\pi_1=\pi_2)$ if and only
if}
\be
\label{eq23}
p(\pi_1) = p(\pi_2) \qquad (\pi_1 = \pi_2) \; .
\ee

\vskip 2mm

{\bf Definition 6.3}. {\it Between two prospects $\pi_1$ and $\pi_2$
belonging to the prospect lattice ${\cal L}$, the prospect $\pi_1$ is
preferred to $\pi_2$ $(\pi_1>\pi_2)$ if and only if}
\be
\label{eq24}
p(\pi_1) > p(\pi_2) \qquad (\pi_1 > \pi_2) \; .
\ee

\vskip 2mm

{\bf Definition 6.4}. {\it A prospect $\pi_*\equiv\sup_n\pi_n$ from the
prospect lattice ${\cal L}$ is optimal if and only if the related prospect
probability is the largest}:
\be
\label{eq25}
p(\pi_*) =\sup_n p(\pi_n) \qquad \Longleftrightarrow \qquad
(\pi_* = \sup_n \pi_n ) \; .
\ee

\vskip 2mm

{\bf Definition 6.5}. {\it The procedure of decision making in QDT
consists of the enumeration of the possible action prospects, the
evaluation of the prospect probabilities, and the choice of the optimal
prospect}.

\section{Interference of Composite Prospects}

The decision procedure described in the previous section, when applied
to composite prospects containing composite actions, results in nontrivial
consequences, often connected to the fact that the probability operators
(\ref{eq17}) for composite prospects correspond to entangling operations
\citep{26,27,28}. Several modes of a composite action can interfere,
leading to the appearance of interference terms. The occurrence of several
modes of an action implies the existence of uncertainty and of the
perception of possible harmful consequences. In contrast, the elementary
prospects (\ref{eq4}) yield no interference. This is because the states of
the elementary prospects are the basic states (\ref{eq8}).

\subsection{Composite prospects}

It is important to distinguish between composite and simple prospects,
since these have different properties related to their structure.

\vskip 2mm

{\bf Definition 7.1} {\it A prospect $\pi_n\in{\cal L}$ is composite if
and only if it is formed by not less than two actions and at least one of
its forming actions is composite}.

\vskip 2mm

The prospect state (\ref{eq13}) can be represented as a superposition
\be
\label{eq26}
| \pi_n > \; = \; \sum_\al \; | \pi_n(e_\al) >
\ee
of the partial weighted states of the elementary prospects,
\be
\label{eq27}
| \pi_n(e_\al) > \; \equiv \; b_n(e_\al)| e_\al > \; .
\ee
The probability of such a weighted elementary prospect, under the
strategic state of mind $|\psi_s>\in{\cal M}$, according to
Eq. (\ref{eq21}), is
\be
\label{eq28}
p_n(e_\al) \equiv | < \pi_n(e_\al) | \psi_s > |^2 \; .
\ee
Taking into account Eqs. (\ref{eq11}) and (\ref{eq27}) gives
\be
\label{eq29}
p_n(e_\al) = | b_n(e_\al) \; c(e_\al) |^2 \; .
\ee
To be classified as a probability, $p_n(e_\al)$ is to be normalized, so
that
\be
\label{eq30}
\sum_{n, \al} \; p_n(e_\al) =  1 \; ,
\ee
which will be always assumed in what follows. The peculiarity of dealing
with composite prospects is that the prospect probability $p(\pi_n)$,
generally, is not a sum of the partial probabilities (\ref{eq29}).

\vskip 2mm

{\bf Proposition 1}. {\it The probability of a composite prospect
$\pi_n\in{\cal L}$, under the strategic state of mind $|\psi_s>\in{\cal M}$,
is the sum
\be
\label{eq31}
p(\pi_n) = \sum_\al \; p_n(e_\al) + q(\pi_n)
\ee
of the partial probabilities of the weighted elementary prospects
(\ref{eq29}) plus the quantum interference term}
\be
\label{eq32}
q(\pi_n) \equiv \sum_{\al\neq\bt}\; b_n^*(e_\al) c(e_\al)
c^*(e_\bt) b_n(e_\bt) \; .
\ee

\vskip 2mm

{\it Proof}: Formula (\ref{eq31}) follows from the definition of the
prospect probability (\ref{eq20}), or (\ref{eq21}), with the
substitutions of Eqs. (\ref{eq11}) and (\ref{eq13}), and with the use
of notations (\ref{eq29}) and (\ref{eq32}).

\vskip 2mm

The use of quantum techniques for defining the prospect probability
yields the appearance in QDT of the effects typical of quantum theory,
such as interference and entanglement. The interference effect in QDT
is analogous to the interference in quantum mechanics \citep{32}.
In QDT, interference happens only for composite prospects, containing
actions represented by several modes. Therefore it involves interferences
between modes.

The entanglement effect may have two different origins. First, the state
of mind $|\psi_s>$ can be expected in general to be entangled. But more
important is that the probability operator (\ref{eq17}), generally, is
an entangling operator, which can create entangled states even from
disentangled ones \citep{26,27,28}. This is somewhat similar to the
generation of entanglement in the transfer of information through quantum
channels \citep{23,24,25}.

There exists several measures of entanglement \citep{23,24,25} and of
entanglement production \citep{26,27,28}. It is important to stress that the
degree of entanglement is correlated with the level of order in the
considered system \citep{28}. This is especially clear if the system order
is classified by means of the {\it order indices} \citep{33,34,35,36}, which
can be introduced for arbitrary operators \citep{37}, and in particular, for
reduced density matrices \citep{38}. Generally speaking, the notions of
order and entanglement are complementary to each other \citep{28}. The more
a system is ordered, the less it is entangled.

In decision theory, the presence of disorder is equivalent to the
existence of uncertainty, and the latter is accompanied by risk \citep{39,12}.
Keeping in mind the above discussion, we can associate the interference
term (\ref{eq32}) with the uncertainty caused by the presence of several
modes representing one action as well as with the perception of potential
harmful consequences accompanying the decision-making process occurring
under uncertainty.

\subsection{Aversion to uncertainty and loss}

The following definitions give an explicit formulation of the {\it
principle of ``aversion to uncertainty and loss''}, which replaces the
principle of risk aversion, as being necessary to understand and explain
the decision making processes of real human beings. The interference term
$q(\pi_n)$, defined in (\ref{eq32}), characterizes the subjective attitude
of a decision maker with respect to the considered prospect $\pi_n$. This
includes the appreciation of the prospect as being potentially more or
less uncertain or harmful, as a result of which the decision maker can look to
the prospect as being more or less attractive. Comparing the terms
in (\ref{eq31}), we can distinguish them as having different meanings.
The first, additive, term, represented by the sum, defines the {\it objective
quantitative} characteristic of the considered prospect, showing its
usefulness. While the second, interference, term $q(\pi_n)$ describes the
{\it subjective qualitative} feature of the prospect, showing how much the
latter is subjectively attractive. This understanding is formalized in the
definitions below.

\vskip 2mm

{\bf Definition 7.2}. {\it The interference term $q(\pi_n)$, defined in
(\ref{eq32}), characterizes the subjective attractiveness, or the quality,
of the prospect $\pi_n$, because of which it can be called the attraction,
or quality, factor}.

\vskip 2mm

{\bf Definition 7.3}. {\it Two prospects, $\pi_1$ and $\pi_2$ from ${\cal L}$,
are equally attractive, or are of equal quality, if and only if
$$
q(\pi_1) \; = \; q(\pi_2) \; .
$$
This implies that they are perceived as equally uncertain or as potentially
equally harmful}.

\vskip 2mm

{\bf Definition 7.4}. {\it Between two prospects, $\pi_1$ and $\pi_2$
from ${\cal L}$, the prospect $\pi_1$ is termed more attractive, or of
better quality, if and only if
$$
q(\pi_1) \; > \; q(\pi_2) \; .
$$
This means that $\pi_2$ is perceived as more uncertain or potentially
more harmful than $\pi_1$}.

\vskip 2mm

From these definitions and formula (\ref{eq31}), it follows that the more
uncertain or potentially more harmful prospect $\pi_2$ possesses a
smaller interference term $q(\pi_1)$, thus, diminishing the probability
$p(\pi_2)$. The stronger perceived uncertainty or potential harm
suppresses the prospect probability. The prospect probability $p(\pi_n)$
is composed of two terms, one defining the usefulness of the prospect and
another characterizing the attractiveness of the latter. In selecting an
optimal prospect, subjects try to avoid potential harm and choose the
prospect that is more useful and, at the same time, more attractive, or
has a subjectively better quality. The probability of choosing this or that
prospect is based on the total value of $p(\pi_n)$, as is defined in
Definitions 6.2 and 6.3.

In order for the previous definitions to be applicable to real decision
processes, it is necessary to specify the notion of uncertainty and of
perceived potential harm by formulating a sufficient condition for one
of the prospects to be treated as more uncertain or more potentially
harmful than another.

\vskip 2mm

{\bf Definition 7.5}. {\it Between two prospects, $\pi_1$ and $\pi_2$ from
${\cal L}$, the prospect $\pi_1$ is said to be more attractive if $\pi_1$
possesses at least one of the following features, as compared to $\pi_2$:

\vskip 2mm

{\rm (a)} getting more certain gain,

{\rm (b)} getting more uncertain loss,

{\rm (c)} being active under certainty,

{\rm (d)} being passive under uncertainty.

In other words, $\pi_2$ is perceived as more uncertain or potentially more
harmful than $\pi_1$. }

\vskip 2mm

{\bf Remark 7.1}. The notions of ``gain" and ``loss" are assumed to have the
standard meaning accepted in the literature on decision making. The same
concerns the notions of ``being active" and ``being passive". The notion
``being active" implies that the decision maker chooses to accomplish an act.
While ``being passive" means that the decision maker restrains from an action.
For instance, in the Hamlet hesitation ``to be or not to be", the first option
``to be" implies activity, while the second possibility ``not to be" means
passivity.

\vskip 2mm

{\bf Remark 7.2}. Note that (b) is the double negation of (a), and (d) is
the double negation of (c). Hence, from the point of view of mathematical
logic \citep{40}, the statements (a) and (b) are equivalent, as are (d)
and (c).

\vskip 2mm

{\bf Remark 7.3}. We are careful to distinguish the concept of ``uncertainty
or perceived potential harm'' from ``risk''. Risk involves the combination
of the uncertainty of a loss and of the severity or amplitude of that loss.
In contrast, uncertainty and perceived potential harm that we consider in
QDT emphasize more the subjective pain that a human subject visualizes in
his/her mind when considering the available options and making a decision.

\subsection{Interference alternation}

The attraction factors enjoy a very important property, without which it
would be impossible to resolve the paradoxes occurring in the classical
decision making.

\vskip 2mm

{\bf Proposition 2}. {\it Let us consider a lattice ${\cal L}$ of
prospects $\pi_n$, with the prospect states $|\pi_n>\; \in {\cal M}$,
under the normalization conditions (\ref{eq22}) and (\ref{eq30}). Then}
\be
\label{eq35}
\sum_n \; q(\pi_n) = 0 \; .
\ee

\vskip 2mm

{\it Proof}: Summing the prospect probabilities (\ref{eq31}) over all
$\pi_n\in{\cal L}$, and using the normalization conditions (\ref{eq22}) and
(\ref{eq30}) yields (\ref{eq35}).

\vskip 2mm

{\bf Remark 7.4}. The interference alternation (\ref{eq35}) shows that some
of the interference terms are positive, while other are negative, so that
the total sum of all these terms is zero. This means that the probability
of prospects with larger uncertainty and/or perceived potential harm will
be suppressed, while that of less uncertain and/or harmful prospects will be
enhanced.

\section{Binary Prospect Lattices}

In the majority of applications, one considers the structures that are
equivalent to a binary lattice ${\cal L}=\{ \pi_1,\pi_2\}$, containing two
composite prospects. In order to address these applications, we specify
below the above mathematical structure to the case of a binary lattice.

\subsection{Binary lattice}

Suppose that $A$ and $X$ are the actions from the action ring ${\cal A}$.
And let $A$ contain just two modes,
\be
\label{eq36}
A = A_1 + A_2 \; .
\ee
While $X\in{\cal A}$ can include $M_2\geq 2$ modes,
\be
\label{eq37}
X = \bigcup_{j=1}^{M_2} X_j \qquad (M_2 \geq 2) \; .
\ee
The mind dimensionality is therefore ${\rm dim}{\cal M}=2 M_2$. The
elementary prospects $A_nX_j$ define the basic states $|A_nX_j>$, with
$n=1,2$ and $j=1,2,\ldots,M_2$. The strategic state of mind (\ref{eq11})
takes the form
\be
\label{eq38}
|\psi_s> \; = \; \sum_{nj}\; c(A_nX_j)| A_n X_j> \; .
\ee

Let the prospect lattice ${\cal L}=\{\pi_n\}$ consist of two prospects
\be
\label{eq39}
\pi_n \equiv A_n X \qquad (n=1,2) \; .
\ee
The prospect states (\ref{eq13}) read as
\be
\label{eq40}
|\pi_n> \; = \; \sum_j \; b(A_nX_j) | A_nX_j> \; .
\ee
The probabilities of elementary prospects (\ref{eq29}) are now
\be
\label{eq41}
p(A_nX_j) = | b(A_nX_j) c(A_nX_j)|^2 \; ,
\ee
with their normalization condition (\ref{eq30}) written as
\be
\label{eq42}
\sum_{nj} \; p(A_n X_j) =  1 \; .
\ee
The quantum interference term (\ref{eq32}) becomes
\be
\label{eq43}
q(\pi_n) = \sum_{i\neq j} \; b^*(A_n X_i) c(A_n X_i)
c^*(A_nX_j) b(A_n X_j) \; .
\ee
Then the prospect probability (\ref{eq21}) yields
\be
\label{eq44}
p(\pi_n) = \sum_j \; p(A_n X_j) + q(\pi_n) \; ,
\ee
in agreement with formula (\ref{eq31}). And the normalization condition
(\ref{eq22}) reduces to
\be
\label{eq45}
p(\pi_1) + p(\pi_2) = 1 \; .
\ee
The theorem on interference alternation (Proposition 2) leads to the
equality
\be
\label{eq46}
q(\pi_1) + q(\pi_2) = 0 \; .
\ee

This case of a binary prospect lattice ${\cal L}=\{\pi_1,\pi_2\}$
explicitly demonstrates that the quantum interference term (\ref{eq43})
is caused by the occurrence of the composite action (\ref{eq37}) making
prospects (\ref{eq39}) composite. The interference between several modes
reflects an uncertainty and/or a perceived potential harm accompanying
the choice of one of the modes representing the composite action.

From the principle of ``aversion to uncertainty and loss'' (Definitions
7.3 and 7.4), one can conclude the following. If the prospects $\pi_1$ and
$\pi_2$ are perceived as equally uncertain and/or potentially harmful, then
the theorem on interference alternation (\ref{eq46}) yields
$q(\pi_1)=q(\pi_2)=0$.

And when one of the prospects, say $\pi_2$, is more uncertain and/or
potentially harmful than $\pi_1$, then, by Definition 7.5 and equality
(\ref{eq46}), one has
\be
\label{eq47}
q(\pi_1) = -q(\pi_2) \; > \; 0 \; , \qquad
q(\pi_1) = |q(\pi_2)| \; > \; 0 \; .
\ee

\subsection{Conditional probabilities}

Let us consider two actions, $A$ and $X$ from the action ring ${\cal A}$,
with the action $A$ being arbitrary and the action $X$ being composite as
in notation (\ref{eq37}). By the definition of the action ring ${\cal A}$,
an action $AX_j$ implies joining two actions $A$ and $X_j$ to be
accomplished together, with the probability $p(AX_j)$. The related
conditional probability $p(A|X_j)$ can be introduced in the standard
manner \citep{41} through the identity
\be
\label{eq48}
p(AX_j) \equiv p(A|X_j)\; p(X_j) \; .
\ee
Here $p(X_j)$ is a prescribed weight of the action $X_j$, satisfying the
conditions
\be
\label{eq49}
\sum_j \; p(X_j) = 1 \; , \qquad
0 \; \leq p(X_j) \; \leq \; 0 \; .
\ee
Interchanging in identity (\ref{eq48}) the actions $A$ and $X_j$, one gets
\be
\label{eq50}
p(X_jA) \equiv p(X_j|A)\; p(A) \; ,
\ee
where $p(A)\equiv p(AX)$ is assumed. The above relations can be formalized
as follows.

\vskip 2mm

{\bf Definition 8.1}. {\it For the actions $A$ and $X$ from the action
ring ${\cal A}$, where $A$ is arbitrary and $X$ is a composite action given
by Eq. (\ref{eq37}), the conditional probability $p(A|X_j)$ of $A$
under condition $X_j$ and the conditional probability $p(X_j|A)$ of $X_j$
under condition $A$ are defined by the equations
\be
\label{eq51}
p(A|X_j) \equiv \frac{p(AX_j)}{p(X_j)} \; , \qquad
p(X_j|A) \equiv \frac{p(X_jA)}{p(AX)} \; ,
\ee
where the weights $p(X_j)$ satisfy normalization (\ref{eq49})}.

\vskip 2mm

It is worth emphasizing that $p(A|X_j)$ does not equal $p(X_j|A)$. Recall
that this is so already in classical probability theory \citep{41}. There is
even less reason to expect their equality in QDT. The relation between
the conditional probabilities (\ref{eq51}) is given by the following
formula below.

\vskip 2mm

{\bf Proposition 3}. {\it The conditional probabilities $p(A|X_j)$ and
$p(X_j|A)$, defined in Eq. (\ref{eq51}), satisfy the relation}
\be
\label{eq52}
p(X_j|A) = \frac{p(X_jA)}{\sum_j p(A|X_j)p(X_j)+q(AX)} \; .
\ee

\vskip 2mm

{\it Proof}: Let us consider the prospect $AX$, with an arbitrary $A$ and
with the composite $X$ given by Eq. (\ref{eq37}). Then, according to Eq.
(\ref{eq44}), the corresponding prospect probability is
$$
p(AX) = \sum_j p(AX_j) + q(AX) \; .
$$
Substituting this into the second of Eqs. (\ref{eq51}) and using Eq.
(\ref{eq48}), we come to relation (\ref{eq52}).

\vskip 2mm

{\bf Remark 8.1}. Formula (\ref{eq52}) is the generalization of the Bayes'
formula of classical probability theory \citep{41}. Equation (\ref{eq52})
reduces to the Bayes formula, provided that there is no interference, when
$q(AX)$ is zero, and that the actions pertain to a field where all actions
are commutative. However, in QDT, the actions belong to a noncommutative
ring ${\cal A}$, so that in general $p(AX_j)$ and $p(X_jA)$ are not equal,
since $AX_j$ is not the same as $X_jA$. As already mentioned, the
noncommutativity of actions is an important feature of QDT.

\subsection{Noncommutativity of actions}

Since the action ring ${\cal A}$ is not commutative with respect to
multiplication, the prospects $AX$ and $XA$ from ${\cal L}$ are generally
different, in the sense that $p(AX)$ does not coincide with $p(XA)$.

\vskip 2mm

{\bf Proposition 4}. {\it Let us consider two prospects $AX$ and $XA$ in
${\cal L}$, with $A\in{\cal A}$ being arbitrary and $X$ being a composite
action given in Eq. (\ref{eq37}). And let at least one of the action modes
$X_j$ be certainly realized under action $A$ so that
\be
\label{eq53}
\sum_j\; p(X_j|A) =  1 \; .
\ee
Then the prospects $AX$ and $XA$ are indifferent $(AX=XA)$ if and only
if there is no mode interference, that is},
\be
\label{eq54}
p(AX) =  p(XA)~,
\ee
{\it if and only if}
\be
\label{eq55}
q(AX) = q(XA) = 0 \; .
\ee

\vskip 2mm

{\it Proof}: For the prospect $XA$, invoking the general procedure, we have
$$
p(XA) = \sum_j \; p(X_jA) + q(XA) \; .
$$
Employing here the definition of the conditional probabilities (\ref{eq51})
gives
$$
p(XA) = \sum_j \; p(X_j|A)\; p(AX) + q(XA) \; .
$$
Using normalization (\ref{eq53}) yields
$$
p(XA) - p(AX) =  q(XA) \; .
$$
Interchanging the actions $A$ and $X$ results in
$$
p(AX) - p(XA) = q(AX) \; .
$$
From the latter two equations, we obtain the statement formalized in Eqs.
(\ref{eq54}) and (\ref{eq55}).

\vskip 2mm

{\bf Remark 8.2}. This theorem emphasizes the intimate relation between the
noncommutativity of actions and the appearance of the interference terms.
The action noncommutativity and the presence of action interference
constitute both characteristic features of QDT.

\section{Explanation of Classical Paradoxes}

A series of paradoxes have been unearthed which cannot find satisfactory
explanations within the framework of classical utility theory. These
paradoxes are well documented in a growing body of empirical evidence.
Numerous related citations can be found in the review articles \citep{42,11}
which provide detailed descriptions of these paradoxes referring to a
voluminous literature. All these studies unambiguously show that decision
makers systematically violate the predictions of classical utility theory,
when decisions are made under uncertainty and risk. There have been
numerous attempts to treat these paradoxes by modifying the expected
utility theory with introducing some more complicated unexpected utility
functionals \citep{11}. However, in addition to being more complicated,
ambiguously defined, and spoiling the nice mathematical properties of the
expected utility, such modifications cannot solve all existing paradoxes in
the classical decision making, as has recently been proved \citep{43}.

The existence of such paradoxes in classical utility theory can be traced
back to the fact that the impact of risk and uncertainty is embodied only
within the choice of the utility function of the decision maker, which is
supposed to fully capture by its functional shape the risk and uncertainty
aversions of the agent. Given a set of competing actions or prospects,
classical utility theory assumes that one can independently and objectively
estimate their corresponding probabilities. Then, the preferred action is
that one which maximizes the expected utility, where the expectation is
performed over all possible scenarios weighted by their corresponding
probabilities. In contrast, QDT takes into account the fact that the
existence of competing actions or prospects in the presence of risk and
uncertainty leads to an entanglement of the probabilities of these
different actions, and therefore to distortions away from pure absolute
objective probabilities. The interference terms and the noncommutativity
of actions, discussed above, describe the fact that the probabilities of
different actions depend on the coexistence of other potential actions
in the mind of the decision maker. From the point of view of QDT, this
single fact is at the origin of the paradoxes in classical utility theory.
Since, as we now show, QDT is able to account qualitatively and
quantitatively for all known classical paradoxes, this suggests that the
axioms of QDT correctly embody at a coarse-grained level the effective
thought processes underlying decision making of real humans.

In the present section, we discuss the main known paradoxes classified as
such by classical utility theory. Naturally, each of these paradoxes can
be represented by an infinite number of variants in different real life
situations. And a large number of such illustrations has been described
in the literature \citep{42,11}. However, from the point of view of
mathematics, there are only a finite number of typical structures. Here,
our aim is specifically to analyze the mathematical structure of these
paradoxes, omitting secondary descriptive details. In so doing, we provide
a general demonstration that no such paradox occurs in the framework of QDT,
allowing one to adapt the same reasoning to the many variants of each
paradox.

\subsection{Compatibility violation}

This paradox, first described by Allais (1953), and now known under his
name, is a choice problem showing an inconsistency of actual observed
choices with the predictions of expected utility theory. It is also often
referred to as the violation of the independence axiom of classical utility
theory. This paradox is that two decisions which are incompatible in the
framework of classical utility theory are nevertheless taken by real
human agents. The mathematical structure of the Allais paradox has been
presented in Sec. 2. Its explanation in the framework of QDT is as
follows.

Let us consider two composite actions
\be
\label{eq56}
A = \bigcup_{n=1}^4 A_n \; , \qquad
X = \bigcup_{j=1}^{\rmM_2} X_j \; ,
\ee
where $\rmM_2\geq 2$. Four action prospects
\be
\label{eq57}
\pi_n \equiv A_n X \qquad (n=1,2,3,4)
\ee
form a quadruple lattice ${\cal L}=\{\pi_n\}$. Suppose that the prospect
ordering is such that $\pi_1$ is strictly preferred to $\pi_2$, hence,
\be
\label{eq58}
p(\pi_1) \; > \; p(\pi_2) \qquad (\pi_1 > \pi_2) \; ,
\ee
with $\pi_2$ being perceived as more uncertain and/or potentially more
harmful than $\pi_1$, that is,
\be
\label{eq59}
q(\pi_2) \; < \; q(\pi_1) \; .
\ee
And let $\pi_3$ be preferred, or indifferent, to $\pi_4$, so that
\be
\label{eq60}
p(\pi_3) \; \geq \; p(\pi_4) \qquad (\pi_3 \geq \pi_4) \; ,
\ee
with the prospects $\pi_3$ and $\pi_4$ being of equal uncertainty and
potential harm,
\be
\label{eq61}
q(\pi_3) = q(\pi_4) \; .
\ee
In addition, one assumes the balance condition
\be
\label{eq62}
p(A_1|X_j) + p(A_3|X_j) = p(A_2|X_j) + p(A_4|X_j) \; .
\ee
In the framework of utility theory, this condition makes incompatible the
decisions associated with the ordering $\pi_1>\pi_2$ and $\pi_3\geq\pi_4$.
In other words, this ordering is  contradictory (see Sec. 2). But this
contradiction does not arise in QDT.

\vskip 2mm

{\bf Proposition 5}. {\it Let a quadruple lattice ${\cal L}=\{\pi_n\}$
of prospects (\ref{eq57}) be ordered so that: (i) $\pi_1>\pi_2$, in the
sense of Eq. (\ref{eq58}), with $\pi_2$ being more uncertain and/or
potentially harmful than $\pi_1$ in the sense of Eq. (\ref{eq59}), and
(ii) $\pi_3\geq\pi_4$ in the sense of Eq. (\ref{eq60}), with $\pi_3$ and
$\pi_4$ being of equal uncertainty and potential perceived harm according
to Eq. (\ref{eq61}). And let the balance condition (\ref{eq62}) be valid.
Then the decisions $\pi_1>\pi_2$ and $\pi_3\geq\pi_4$ are compatible,
provided that}
\be
\label{eq63}
0\; \leq \; \sum_j \; [ p(A_2X_j) - p(A_1X_j) ] \; < \;
q(\pi_1) - q(\pi_2) \; .
\ee

\vskip 2mm

{\it Proof}: The probabilities of the prospects (\ref{eq57}) read as
\be
\label{eq64}
p(\pi_n) = \sum_j \; p(A_n|X_j)\; p(X_j) + q(\pi_n) \; .
\ee
Since $\pi_1>\pi_2$, in the sense of inequality (\ref{eq58}), we have
\be
\label{eq65}
\sum_j \; [ p(A_1|X_j) - p(A_2|X_j) ]\; p(X_j) \; > \;
q(\pi_2) - q(\pi_1) \; .
\ee
Because of $\pi_3\geq\pi_4$, according to Eq. (\ref{eq60}), we get
\be
\label{eq66}
\sum_j \; [ p(A_3|X_j) - p(A_4|X_j) ]\; p(X_j) \; > \;
q(\pi_4) - q(\pi_3) \; .
\ee
Equation (\ref{eq65}) can be rewritten as
\be
\label{eq67}
\sum_j \; [ p(A_2X_j) - p(A_1X_j) ]  \; < \;
q(\pi_1) - q(\pi_2) \; .
\ee
And Eq. (\ref{eq66}), invoking the balance condition (\ref{eq62}),
transforms into
\be
\label{eq68}
\sum_j \; [ p(A_1X_j) - p(A_2X_j) ]  \; \leq \;
q(\pi_3) - q(\pi_4) \; .
\ee
Taking into account condition (\ref{eq61}) reduces Eq. (\ref{eq68}) to
\be
\label{eq69}
\sum_j \; [ p(A_2X_j) - p(A_1X_j) ]  \; \geq \; 0 \; .
\ee
Combining Eqs. (\ref{eq67}) and (\ref{eq69}) yields the desired result
(\ref{eq63}).

\vskip 2mm

{\bf Remark 9.1}. In the classical utility theory, there are no interference
terms. The reduction of QDT to classical utility theory is obtained by
setting the terms $q(\pi_n)$ to zero. Then, Eq. (\ref{eq63}) becomes a
contradiction, which is nothing but the Allais paradox. In QDT, there is
no contradiction (and no paradox) since the right-hand side of Eq.
(\ref{eq63}) is positive according to condition (\ref{eq59}).

\subsection{Independence paradox}

As has been already mentioned, the Allais paradox not only reveals that
real human beings can select decisions which are mutually incompatible
within the standard expected utility theory, but it also shows that the
independence axiom of utility theory is violated by real humans. The
independence axiom stipulates that, if $\pi_1 > \pi_2$ and $\pi_3\geq\pi_4$,
then $\pi_1+\pi_3>\pi_2+\pi_4$. The mathematics of how this independence
axiom breaks down is explained in Sec. 2. In contrast, within QDT, the
violation of the independence-axiom does not occur. Instead, we can state
the following independence theorem.

\vskip 2mm

{\bf Proposition 6}. {\it Let a lattice ${\cal L}=\{\pi_n\}$
of six prospects  be ordered so that $\pi_1 > \pi_2$, with $\pi_1$
being perceived as more uncertain and/or potentially more harmful than
$\pi_2$, and $\pi_3\geq\pi_4$, with $\pi_3$ and $\pi_4$ being of equal
uncertainty and potential harm. And let $\pi_2+\pi_4$ be more uncertain
and/or more potentially harmful than or at the same perceived level as
$\pi_1+\pi_3$, so that
\be
\label{eq70}
q(\pi_2+\pi_4) \; \leq \; q(\pi_1+\pi_3)~.
\ee
Then the prospect $\pi_1+\pi_3$ is strictly preferred to $\pi_2+\pi_4$,
that is},
\be
\label{eq71}
p(\pi_1+\pi_3) \; > \; p(\pi_2+\pi_4) \; .
\ee

\vskip 2mm

{\it Proof}: By employing the definition of the prospect
probabilities defined above, we have
$$
p(\pi_1+\pi_3) \equiv | < \pi_1 + \pi_3|\psi_s> |^2 \; , \qquad
p(\pi_2+\pi_4) \equiv | < \pi_2 + \pi_4|\psi_s> |^2 \; ,
$$
for a given state of mind $|\psi_s> $. Straightforward calculations give
$$
p(\pi_1 + \pi_3) = p(\pi_1) + p(\pi_3) + q(\pi_1 + \pi_3) \; ,
\qquad p(\pi_2 + \pi_4) = p(\pi_2) + p(\pi_4) +
q(\pi_2 + \pi_4) \; .
$$
Using inequalities (\ref{eq58}), (\ref{eq60}), and (\ref{eq70}) yields
the above inequality (\ref{eq71}).

\vskip 2mm

{\bf Remark 9.2}. Let us stress that invoking the notion of uncertainty
and/or perceived potential harm, which is rigorously related to the
occurrence of the interference terms, makes Proposition 6 a theorem
rather than an assumption.

\subsection{Utility failure}

Another well-known anomaly in the use of utility theory to account for
real human decisions is called the {\it Ellsberg paradox} \citep{45}. It
states that, in some cases, no utility function can be defined at all, so
that utility theory fails. The mathematical structure of the Ellsberg
paradox is described in Sec. 2. As we show below, such a paradox does
not arise in QDT.

Let us consider two composite actions
\be
\label{eq72}
A = A_1 + A_2 \; , \qquad X= \bigcup_{j=1}^{M_2} X_j \; ,
\ee
where $M_2\geq 2$. The case of interest is the binary lattice
${\cal L}=\{\pi_n\}$ of the prospects
\be
\label{eq73}
\pi_n \equiv A_n X \qquad (n=1,2) \; .
\ee
The Ellsberg paradox is characterized by the equivalence condition
\be
\label{eq74}
p(A_1|X_j) = p(A_2|X_j) \qquad (\forall j) \; .
\ee
The prospects $\pi_1$ and $\pi_2$ are distinguished by the fact that
$p(A_1|X_j)$ is given explicitly, while $p(A_2|X_j)$ is not explicitly
available, but its existence is just stated. While the structure of the
decision making process is such that (\ref{eq74}) holds true, its
formulation adds what has been termed an ``ambiguity'' for one of the
prospects ($\pi_2$). Within QDT, this ambiguity is simply taken into
account by saying that $\pi_2$ is perceived as more uncertain than
$\pi_1$ which, according to Definition 7.5, implies
\be
\label{eq75}
q(\pi_2) \; < \; q(\pi_1)  \; .
\ee
Human subjects are found to decide in favor of the less uncertain prospect
$\pi_1$, which is usually preferred to $\pi_2$. But, as explained in
Sec. 2, the option $\pi_1 > \pi_2$, under condition (\ref{eq74}), cannot
be correct with any utility function. In contrast, within QDT, the
choice $\pi_1 > \pi_2$ is not merely admissible but is compulsory, given
the structure of the problem. The prediction of QDT for this problem can be
stated under the following proposition.

\vskip 2mm

{\bf Proposition 7}. {\it Let us consider the binary lattice ${\cal L}=
\{\pi_n\}$ of prospects (\ref{eq73}), supplemented by condition (\ref{eq74}).
And let $\pi_2$ be perceived as more uncertain than $\pi_1$. Then $\pi_1$ is
preferred to $\pi_2$, that is,}
\be
\label{eq76}
p(\pi_1) \; > \; p(\pi_2) \; .
\ee

\vskip 2mm

{\it Proof}: Following the mathematical structure of QDT, we have
$$
p(\pi_1) - p(\pi_2) = \sum_j \; \left [ p(A_1|X_j) - p(A_2|X_j)
\right ] \; p(X_j) \; + \; q(\pi_1) - q(\pi_2) \; ,
$$
Invoking condition (\ref{eq74}) yields
$$
p(\pi_1) - p(\pi_2) = q(\pi_1) - q(\pi_2) \; .
$$
Taking into account that $\pi_2$ is more uncertain than $\pi_1$, in the
sense of Eq. (\ref{eq75}), we obtain inequality (\ref{eq76}) telling
us that $\pi_1 > \pi_2$ ($\pi_1$ is preferred to $\pi_2$).

\vskip 2mm

{\bf Remark 9.3}. The reduction of QDT to classical utility theory is
obtained, as usual, by putting the interference terms to zero. In this case,
$p(\pi_1)$ is equal to $p(\pi_2)$, which is incompatible with the choice
$\pi_1 > \pi_2$ performed by most real human beings, retrieving the
Ellsberg paradox.

\subsection{Inversion paradox}

A large set of paradoxes found when applying classical utility theory
to the decision making of real human beings are related to the unexpected
inversion of choice, when decisions are made in the presence of uncertainty.
In other words, the ordering or preference of competing choices according
to classical utility theory is reversed by human beings. For this
literature, we refer to the numerous citations found in \cite{42} and
\cite{11}. This anomaly is sometimes called the Rabin paradox \citep{46}.

Let us consider two composite actions
\be
\label{eq77}
A = \bigcup_{n=1}^{M_1} A_n \; , \qquad
X = \bigcup_{j=1}^{M_2} X_j \; ,
\ee
where $M_1\geq 2$ and $M_2\geq 2$. The prospect lattice
${\cal L}=\{\pi_n\}$ is composed of $M_1$ prospects
\be
\label{eq78}
\pi_n \equiv A_n X \qquad (n=1,2,\ldots, M_1) \; .
\ee
Suppose that the partial probabilities obey the majorization condition
\be
\label{eq79}
\sum_j \; p(A_1X_j) \; \geq \; \sum_j \; p(A_2 X_j) \; .
\ee
From the point of view of classical utility theory, under condition
(\ref{eq79}), the prospect $\pi_1$ should be preferred or equivalent to
$\pi_2$. However, subjects often decide in favor of $\pi_2$, when $\pi_1$
is perceived as more uncertain than $\pi_2$, which contradicts utility
theory. But this contradiction is removed in QDT, as formulated by the
following proposition.

\vskip 2mm

{\bf Proposition 8}. {\it Let us consider a lattice ${\cal L}=\{\pi_n\}$
of prospects (\ref{eq78}). And let condition (\ref{eq79}) be valid.
Nevertheless, $\pi_2$ is preferred to $\pi_1$, that is $p(\pi_2) > p(\pi_1)$,
when $\pi_1$ is more uncertain than $\pi_2$, such that}
\be
\label{eq80}
q(\pi_2) - q(\pi_1) \; > \; \sum_j \;
\left [ p(A_1X_j) - p(A_2X_j) \right ] \; .
\ee

\vskip 2mm

{\it Proof}: For the difference of the prospect probabilities, we have
$$
p(\pi_1) - p(\pi_2) = \sum_j \; \left [ p(A_1X_j) - p(A_2X_j)
\right ] \; + \; q(\pi_1) - q(\pi_2)\; .
$$
This, together with condition (\ref{eq80}), gives $p(\pi_2) > p(\pi_1)$,
which means that $\pi_2$ is preferred to $\pi_1$.

\vskip 2mm

{\bf Remark 9.4}. Within classical utility theory, characterized by the
absence of interference terms, condition (\ref{eq80}) would result in the
inverse conclusion that $\pi_1$ is preferred to $\pi_2$. The inversion
paradox occurs in situations where subjects would nevertheless opt for
$\pi_2 > \pi_1$.

\subsection{Invariance violation}

This paradox was described by Kahneman and Tversky (1979), who pointed out
that in some cases utility theory yields the same expected utility outcomes
for several prospects, while subjects clearly prefer some prospects to
others. The mathematical structure of the Kahneman-Tversky paradox is
explained in Sec. 2.

One considers four composite prospects, as in Eq. (\ref{eq78}), under the
invariance condition
\be
\label{eq81}
\sum_j \; p(A_nX_j) = const \qquad (n=1,2,3,4) \; .
\ee
Since condition (\ref{eq81}) leads to the invariance of the expected
utility functions with respect to permutations or replacements among
these four prospects, if subjects were basing their decision according
to the recipes of classical utility theory, they should be indifferent
with respect to the four choices. However, real human subjects demonstrate
evident preference for some of the prospects, when the choices are
accompanied by uncertainty. Kahneman and Tversky characterize the prospect
$\pi_1$ as yielding more uncertain gain than $\pi_2$. And the prospect
$\pi_4$ is defined as resulting in a more certain loss than $\pi_3$, that
is, $\pi_4$ is more potentially harmful than $\pi_3$. Summarizing, one has
\be
\label{eq82}
q(\pi_1) \; < \; q(\pi_2) \; , \qquad q(\pi_4) \; < \; q(\pi_3) \; .
\ee
Under these conditions, there is no paradox remaining within QDT, as
shown by the following proposition.

\vskip 2mm

{\bf Proposition 9}. {\it Let us consider a quadruple lattice ${\cal L}=
\{\pi_n\}$ of the composite prospects (\ref{eq78}), under the invariance
condition (\ref{eq81}) and uncertainty conditions (\ref{eq82}). Then
$\pi_2$ is preferred to $\pi_1$ and $\pi_3$ is preferred to $\pi_4$, so
that}
\be
\label{eq83}
p(\pi_2) \; > \; p(\pi_1) \; , \qquad p(\pi_3) \; > \; p(\pi_4) \; .
\ee

\vskip 2mm

{\it Proof}: It is easy to notice that the Kahneman-Tversky paradox
is nothing but a slightly complicated version of the Ellsberg paradox.
The Kahneman-Tversky paradox can be treated as a particular case of the
inversion paradox. Therefore the proof of Eqs. (\ref{eq83}) is the
same as in Propositions 7 and 8.

\vskip 2mm

{\bf Remark 9.5}. In classical utility theory, with no interference
terms, all prospect probabilities would be the same for all prospects,
contradicting inequalities (\ref{eq83}). This is the meaning of the
Kahneman-Tversky paradox of invariance violation (see Sec. 2).

\subsection{Certainty effects}

A number of paradoxes, related to the description of prospects as more
or less certain, and robustly violating expected utility theory, are
collectively called certainty effects. Typical examples of such paradoxes
have already been discussed, such as the Allais paradox, the Ellsberg
paradox, and the Kahneman-Tversky paradox, when real human subjects prefer
a less uncertain prospect notwithstanding the fact that it has a smaller
expected utility. The mathematical structure of such paradoxes, analyzed
in Sec. 2, demonstrates that these paradoxes are associated with
difficulties in discriminating between different reward values.

In addition to these direct certainty effects, there exist reversed
certainty effects \citep{48}, when subjects prefer a more uncertain
prospect. This happens when rewards are easily discriminated, being
evidently different. The reversed certainty effects can also be explained
in the framework of QDT.

\vskip 2mm

{\bf Proposition 10}. {\it Let us consider a lattice ${\cal L}=\{\pi_n\}$
of prospects defined in Eq. (\ref{eq78}). Suppose that a prospect $\pi_1$
is more uncertain or more potentially harmful that $\pi_2$, in the sense
that $q(\pi_1) < q(\pi_2)$. Despite the fact that the prospect $\pi_1$ is
more uncertain, it is preferred to $\pi_2$ when and only when}
\be
\label{eq84}
\sum_j \; \left [ p(A_1 X_j) - p(A_2 X_j) \right ] \; > \;
q(\pi_2) - q(\pi_1) \; > \; 0 \; .
\ee

\vskip 2mm

{\it Proof}: Writing the difference $p(\pi_1)-p(\pi_2)$, as in
Proposition 8, and requiring that $p(\pi_1)$ be larger than $p(\pi_2)$
yields condition (\ref{eq84}).

\vskip 2mm

{\bf Remark 9.6}. When the left-hand side of Eq. (\ref{eq84}) is close to
zero, this inequality cannot be true in the presence of dissimilar
uncertainties between the two prospects. Instead, inequality (\ref{eq80})
holds, since, by assumption, $q(\pi_1)<q(\pi_2)$. That is, we retrieve the
direct certainty effect. But when the left-hand side of Eq. (\ref{eq84}) is
sufficiently large, this inequality may become valid, while Eq. (\ref{eq80})
looses its validity. This means that the reversed certainty effect holds
true. These conclusions confirm that the direct certainty effects emerge
when it is difficult to discriminate between rewards, whereas the reversed
certainty effects arise when the discrimination of rewards is easy, in
agreement with empirical observations on decision making under uncertainty
by human subjects \citep{48}.

\subsection{Disjunction effect}

The disjunction effect is the violation of the Savage ``sure-thing
principle" \citep{49}. According to this principle, if an alternative
$A_1$ is preferred to an alternative $A_2$, when an event $X_1$ occurs,
and it is also preferred to $A_2$, when an event $X_2$ occurs, then
$A_1$ should be preferred to $A_2$, when it is not known which of the
events, either $X_1$ or $X_2$, has happened (the so-called``sure-thing''
principle). In real life, human subjects often violate the sure-thing
principle, this violation being called the disjunction effect. In the
literature, there are hundreds of examples of concrete realizations of the
disjunction effect \citep{42,50,11} and references therein), all of them
having the same mathematical structure.

One considers a binary lattice ${\cal L}=\{\pi_n\}$ of prospects
\be
\label{eq85}
\pi_n \equiv A_n X \; , \qquad X = \bigcup_{j=1}^{M_2} X_j \; ,
\ee
where $n=1,2$ and $M_2\geq 2$. The majorization condition
\be
\label{eq86}
p(A_1X_j) \; > \; p(A_2 X_j) \qquad (\forall j)
\ee
is assumed. The Savage ``sure-thing principle" \citep{49} states that, under
condition (\ref{eq86}), the prospect $\pi_1$ must be preferred
to $\pi_2$. This, however, does not happen when $\pi_1$ is sufficiently
more uncertain than $\pi_2$, a situation which leads subjects to prefer
$\pi_2$ over $\pi_1$, in blatant contradiction with the Savage principle.
The following proposition shows how this contradiction is removed within
QDT.

\vskip 2mm

{\bf Proposition 11}. {\it Let us consider a binary lattice ${\cal L}=
\{\pi_1,\pi_2\}$, with $\pi_1$ being more uncertain than $\pi_2$, i.e.,
$q(\pi_1) < q(\pi_2)$. And let the majorization condition (\ref{eq86}) be
valid. Notwithstanding this majorization, the prospect $\pi_2$ is preferred
to $\pi_1$, so that $p(\pi_2) > p(\pi_1)$, provided that}
\be
\label{eq87}
q(\pi_2) \; > \; \frac{1}{2} \; \sum_j \;
\left [ p(A_1X_j) - p(A_2 X_j) \right ] \; .
\ee

\vskip 2mm

{\it Proof}:  We notice that the majorization condition (\ref{eq86})
is a sufficient condition for inequality (\ref{eq79}) to hold, so
that the disjunction effect is a particular case of the inversion
paradox. Following the same proof as in Proposition 8, we get inequality
(\ref{eq80}) as a condition for $\pi_2 > \pi_1$. Using the theorem on
interference alternation (Proposition 2), which for a binary lattice takes
the form of Eqs. (\ref{eq47}), we come to condition (\ref{eq87}) under
which $\pi_2 > \pi_1$.

\vskip 2mm

{\bf Remark 9.7}. Equation (\ref{eq87}) clearly shows the origin of the
disjunction effect, contradicting the Savage ``sure-thing principle". In
classical utility theory, where the interference term $q(\pi_2)$
is zero, the left-hand side of Eq. (\ref{eq87}) is zero, while the
right-hand side, owing to condition (\ref{eq86}), is positive, leading to
a contradiction. It is easy to check numerically that condition
(\ref{eq87}) always holds when the disjunction effect is observed
empirically \citep{42,50,11}.

\subsection{Conjunction fallacy}

The conjunction rule of classical probability theory requires that the
probability of the conjunction of two events cannot be larger than any of
the probabilities of the separate events forming the conjunction. However,
again, when decisions are made under uncertainty, subjects often reliably
violate the conjunction rule, which is termed the conjunction fallacy
\citep{42}. There exist numerous particular examples of this fallacy
\citep{42,Yat86,11}.

To describe the mathematical structure of the problem, let us consider
two composite actions $A$ and $X$ and a lattice ${\cal L}=\{\pi_n\}$ of
the prospects
\be
\label{eq88}
\pi_n \equiv A_n X \; , \qquad A = \bigcup_{n=1}^{M_1} A_n \; ,
\qquad X = \bigcup_{j=1}^{M_2} X_2 \; ,
\ee
with $M_1\geq 2$ and $M_2\geq 2$. In applications, one usually takes
$M_1$ and $M_2$ equal to two. But the mathematical structure of the problem
does not change for arbitrary $M_1\geq 2$ and $M_2\geq 2$. In classical
probability theory, $p(\pi_n)$ can never be smaller than any $p(A_nX_j)$
for arbitrary $n$ and $j$. This is why, when it happens that there is at
least one pair of $n=n_0$ and $j=j_0$, such that subjects support the
following inequality,
\be
\label{eq89}
p(\pi_{n_0}) \; < \; p(A_{n_0} X_{j_0} ) \; ,
\ee
this is classified as the conjunction fallacy. There can exist several
$n$ and $j$ for which the conjunction fallacy (\ref{eq89}) occurs. The
most general case is when there is at least one $n_0$, for which
condition (\ref{eq89}) holds true, with $j_0$ corresponding to the
largest partial probability, such that
\be
\label{eq90}
 p(A_{n_0} X_{j_0} ) \equiv \sup_j  p(A_{n_0} X_j ) \; .
\ee
In QDT, the conjunction-fallacy paradox finds a natural explanation, if
decisions are made in the presence of uncertainty.

\vskip 2mm

{\bf Proposition 12}. {\it Let us consider a lattice ${\cal L}=\{\pi_n\}$
of prospects defined in Eqs. (\ref{eq88}). The conjunction fallacy, in the
sense of inequality (\ref{eq89}), happens when and only when there
exists at least one uncertain prospect $\pi_{n_0}\in{\cal L}$, for which}
\be
\label{eq91}
q(\pi_{n_0}) \; < \; - \sum_{j(\neq j_0)} \; p(A_{n_0} X_j ) \; .
\ee

\vskip 2mm

{\it Proof}: From the difference
$$
p(\pi_{n_0}) - p(A_{n_0} X_{j_0} ) = \sum_j \; p(A_{n_0} X_j ) \; - \;
p(A_{n_0} X_{j_0} ) + q(\pi_{n_0})
$$
it follows that inequality (\ref{eq91}) is the necessary and sufficient
condition for Eq. (\ref{eq89}) to be true.

\vskip 2mm

{\bf Remark 9.8}. As has been stressed above, the reduction to classical
probability theory can be done by setting all interference terms
$q(\pi_n)$ to zero. But then inequality (\ref{eq91}) could never be true,
showing that the conjunction fallacy is a real paradox in the framework of
the classical theory. In QDT, as we have numerically checked in a number
of concrete examples, condition (\ref{eq91}) always holds in the situations
where the conjunction fallacy has been observed with human subjects
\citep{42,11}.

\subsection{Isolation effect}

The isolation effect is a common name for a large variety of phenomena
known under different guises, such as the focusing effect, the availability
effect, the salience effect, the framing effect, or elicitation
effect \citep{51}. The essence of all these phenomena is that, when choosing
among several alternatives, subjects are more inclined towards the certain
benefits and at the same time they prefer the prospects with more uncertain
losses. For instance, there exists a well documented tendency to prefer
hidden to transparent taxes \citep{51}. Subjects tend to prefer what is
related to more uncertain loss, while not being necessarily optimal from the
point of view of classical utility theory.

According to Definition 7.5, an alternative with more certain loss is more
repulsive than that one with more uncertain loss. In other words,
subjects simply prefer alternatives that are less painful to them, when
they choose those having more uncertain losses. Using this view point, the
origin of the isolation effect can be easily understood in the framework
of QDT.

Let us consider a lattice ${\cal L}=\{\pi_n\}$ of prospects (\ref{eq88}).
And let the actions $A_j$ be enumerated so that
\be
\label{eq92}
\sum_j \; p(A_1X_j) = \sup_n \sum_j \; p(A_n X_j) \; .
\ee
Then, according to classical utility theory, the prospect $\pi_1$ should
be preferred to all other $\pi_n$, with $n\neq 1$. In reality, however,
this does not happen \citep{51}, when $\pi_1$ is more uncertain or
potentially more harmful. Subjects choose another prospect $\pi_n$ from
${\cal L}$, with $n\neq 1$, which is less uncertain or harmful than $\pi_1$.

\vskip 2mm

{\bf Proposition 13}. {\it For a lattice ${\cal L}=\{\pi_n\}$ of
prospects (\ref{eq88}), under condition (\ref{eq92}), there exists a
prospect $\pi_n$, with $n\neq 1$, which is preferred to $\pi_1$, when
and only when the prospect $\pi_1$ is so uncertain or harmful that}
\be
\label{eq93}
q(\pi_1) \; < \; q(\pi_n) \; + \; \sum_j \; p(A_nX_j) \; - \;
\sup_n \sum_j \; p(A_n X_j) \; .
\ee

\vskip 2mm

{\it Proof}: Equation (\ref{eq92}) is a slight generalization of
the majorization condition (\ref{eq79}) in the inversion paradox.
Therefore the proof of inequality (\ref{eq93}) is the same as in
Proposition 8. If Eq. (\ref{eq92}) is strengthened by the majorization
condition (\ref{eq86}), then the isolation effect reduces to the
disjunction effect and can be treated as in Proposition 11.

\vskip 2mm

{\bf Remark 9.9}. In classical utility theory, where there is no decision
interference, all $q(\pi_n)$ are vanishing. In this case, the left-hand
side of inequality (\ref{eq93}) is zero, while the right-hand side is
negative, leading to a contradiction. The isolation effect
is thus a paradox in the framework of  classical utility theory, which is
absent in QDT.

\subsection{Combined paradoxes}

It may happen that several paradoxes among those considered above
occur simultaneously. In the literature, one usually studies each paradox
separately. This suggests to consider different experimental situations in
which several paradoxes occur simultaneously in order to test our QDT
through novel predictions in context not yet explored. To prove
theoretically that such a situation can really happen, we analyze below the
conditions under which the conjunction fallacy could coexist with the
inversion paradox, and therefore, with the disjunction effect which is a
particular case of the inversion paradox.

We consider two actions $A=A_1+A_2$ and $X=X_1+X_2$ from the action ring
${\cal A}$. And let the prospect $\pi_n$ be formed as in Eq. (\ref{eq88}),
with $M_1=M_2=2$. We are thus dealing with the binary lattice
\be
\label{eq94}
{\cal L} = \{ \pi_n \equiv A_n X : \; n=1,2 \} \; .
\ee
Suppose that the $X_j$'s are enumerated so that
\be
\label{eq95}
p(A_1 X_1) \equiv \sup_j p(A_1 X_j) \qquad (j=1,2) \; .
\ee
This is just a notation, which simplifies the following formulas. In this
notation, the conjunction-fallacy condition (\ref{eq89}), taking account
of Eq. (\ref{eq90}), simplifies to
\be
\label{eq96}
p(\pi_1) \; < \; p(A_1X_1) \; .
\ee
Depending on the value of $p(A_1X_1)$, three different situations can be
found.

\vskip 2mm

{\bf Proposition 14}. {\it Let us consider the binary lattice (\ref{eq94}),
with notation (\ref{eq95}).  And assume that
\be
\label{eq97}
p(A_1X_1) \; < \; \frac{1}{2} \; .
\ee
Then the conjunction-fallacy condition (\ref{eq96}) makes the prospect
$\pi_2$ preferred to $\pi_1$, so that $p(\pi_1) < p(\pi_2)$}.

\vskip 2mm

{\it Proof}. Suppose that the conjunction-fallacy condition (\ref{eq96})
holds, which is a particular case of Eq. (\ref{eq89}). According to
Proposition 12, the necessary and sufficient condition for the validity
of Eq. (\ref{eq89}) is inequality (\ref{eq91}). The latter, with notation
(\ref{eq95}), reads as
\be
\label{eq98}
q(\pi_1) \; < \; - p(A_1X_2) \; .
\ee
Keeping in mind the interference alternation theorem (Proposition 2),
which for a binary lattice takes the form of Eqs. (\ref{eq47}), inequality
(\ref{eq98}) can be rewritten as
\be
\label{eq99}
q(\pi_2) \; > \; p(A_1X_2) \; .
\ee
For the right-hand side of Eq. (\ref{eq99}), from condition (\ref{eq97}),
we have
\be
\label{eq100}
p(A_1X_2) \; > \; \sum_j \; p(A_1 X_j) \; - \; \frac{1}{2} \; .
\ee
Using the normalization condition (\ref{eq42}) gives
\be
\label{eq101}
\sum_j \; p(A_1 X_j) \; - \; \frac{1}{2} \; = \; \frac{1}{2} \sum_j \;
\left [ p(A_1 X_j)  - p(A_2 X_j) \right ] \; .
\ee
In this way, inequality (\ref{eq98}) acquires the form
\be
\label{eq102}
q(\pi_2) \; > \; p(A_1X_2) \; > \; \frac{1}{2} \sum_j \;
\left [ p(A_1 X_j)  - p(A_2 X_j) \right ] \; .
\ee
This shows that Eq. (\ref{eq87}) is valid, which, in the case of a binary
lattice, is the necessary and sufficient condition for $\pi_2 > \pi_1$
($\pi_2$ is preferred to $\pi_1$).

\vskip 2mm

{\bf Proposition 15}. {\it Let us suppose that for the binary lattice
(\ref{eq94}), with notation (\ref{eq95}), the inequality
\be
\label{eq103}
p(A_1 X_1) \; > \; \frac{1}{2}
\ee
holds. Then the assumption that $\pi_2$ is preferred to $\pi_1$,
in the sense that $p(\pi_1) < p(\pi_2)$, results in the conjunction-fallacy
condition (\ref{eq96})}.

\vskip 2mm

{\it Proof}: For a binary lattice, the necessary and sufficient
condition for $p(\pi_1) < p(\pi_2)$ is inequality (\ref{eq87}). Employing
Eq. (\ref{eq101}) gives
\be
\label{eq104}
q(\pi_2) \; > \; \sum_j \; p(A_1 X_j) \; - \; \frac{1}{2} \; .
\ee
Under condition (\ref{eq103}), one has
\be
\label{eq105}
\sum_j \; p(A_1 X_j) \; - \; \frac{1}{2} \; > \; p(A_1 X_2) \; .
\ee
Combining Eqs. (\ref{eq104}) and (\ref{eq105}) yields Eq. (\ref{eq99}),
which is equivalent to inequality (\ref{eq98}). The latter is the
necessary and sufficient condition for the conjunction-fallacy Eq.
(\ref{eq96}).

\vskip 2mm

{\bf Proposition 16}. {\it Let us assume that for the binary lattice
(\ref{eq94}), with notation (\ref{eq95}), the equality
\be
\label{eq106}
p(A_1 X_1) = \frac{1}{2}
\ee
is satisfied. Then the conjunction-fallacy condition (\ref{eq96})
is the necessary and sufficient condition for $\pi_2$ to be preferred
to $\pi_1$, so that $p(\pi_1) < p(\pi_2)$}.

\vskip 2mm

{\it Proof}: The proof is the same as in Propositions 14 and 15,
except that inequalities (\ref{eq100}) and (\ref{eq105}), under equality
(\ref{eq106}), are replaced by the equality
\be
\label{eq107}
p(A_1 X_2) = \sum_j\; p(A_1 X_j) \; - \; \frac{1}{2} \; .
\ee

\vskip 2mm

{\bf Remark 9.10}. The proofs of Propositions 14, 15, and 16 do not depend
on the relation between $p(A_1X_j)$ and $p(A_2X_j)$. Therefore, if in
addition to the assumptions underlying these propositions, we invoke the
majorization condition (\ref{eq79}), then the above propositions describe
the interconnection between the conjunction fallacy and the inversion
paradox. And if we add the majorization conditions (\ref{eq86}), then we
obtain the relation between the conjunction fallacy and the disjunction
effect. The simultaneous occurrence of the latter two effects has not been
analyzed experimentally, as far as we are aware. But, as we have proved,
it may happen. From the analysis of the available experimental data on
the conjunction fallacy \citep{42,11}, we find that the latter is usually
characterized by condition (\ref{eq97}). Thus, we predict that the
conjunction fallacy should be accompanied by the disjunction effect.

\section{Outlook}

We have presented a detailed description of Quantum Decision Theory
(QDT), whose ideological source can be found in the insightful writings
of Niels Bohr on the qualitative interpretations of measurements in
quantum theory \citep{2,3,4}. Bohr advocated the feasibility of describing
psychological processes, such as decision making, by means of
mathematical techniques of quantum theory. Developing the theory of
measurement, von Neumann (1955) noted that the projection operators can be
treated as statements about events, with the operator expectations
providing the event probabilities. The von Neumann theory of quantum
measurement describes the situation when an observer accomplishes
measurements on a passive quantum system \citep{6,7,8}. We have extended
the von Neumann theory in several aspects summarized below.

\vskip 2mm

(1) The principal novelty of our approach viewed from the vantage of
measurement theory is the possibility of applying it to {\it active}
systems making decisions. This is achieved by introducing a specific state,
the strategic state of mind, characterizing the considered decision maker.

\vskip 2mm

(2) We have specified the basic techniques of QDT so that they could be
applicable to real decision processes. In particular, the manifold
of intended actions is defined as a noncommutative ring, since
noncommutativity is a typical property that captures accurately what we
believe is an essential property of human decision making. The set of
action prospects is characterized as a complete lattice.

\vskip 2mm

(3) The point of fundamental importance in our approach is that
the action prospects are described as composite objects, formed by
composite actions. The composite structure of prospects, together with
the entangling properties of probability operators, result in the
appearance of decision interferences, which take into account the
uncertainties and repulsion to potential harmful consequences associated
with the decision procedure.

\vskip 2mm

(4) The noncommutativity of prospects and their interference are shown
to be intimately connected, and they arise simultaneously.

\vskip 2mm

(5) The interference of prospects is proved to exhibit the property of
alternation, leading to the suppression of the probabilities of more
uncertain or potentially more harmful prospects in favor of those of less
uncertain or less harmful prospects which are comparatively enhanced.

\vskip 2mm

(6) We have demonstrated that practically all known anomalies and paradoxes
documented in the context of classical utility theory are reducible to just
a few mathematical archetypes,  all of which finding straightforward
explanations in the framework of QDT.

\vskip 2mm

(7) We predict that some of the analyzed effects can occur together. For
instance, the conjunction fallacy is found to be a sufficient condition
for the disjunction effect.

\vskip 2mm

The major novel results of the present paper is the mathematical
development of the decision theory, based on quantum rules, and the proof
that all known paradoxes of classical decision theory find a natural
explanation in the frame of this quantum decision theory.

\vskip 5mm

{\bf Acknowledgements}

\vskip 2mm

We appreciate financial support from the ETH Competence Center for Coping
with Crises in Socio-Economic Systems (CCSS) through the ETH Research
Grant CH1-01-08-2. We are also grateful to the participants of the CCSS
Seminars for their remarks. One of the authors (V.I.Y.) is grateful to
P.A. Benioff for useful correspondence and advice. We appreciate helpful
discussions with E.P. Yukalova.

\end{document}